\def\year{2022}\relax
\documentclass[letterpaper]{article} 
\usepackage{aaai22}  
\usepackage{times}  
\usepackage{helvet}  
\usepackage{courier}  
\usepackage[hyphens]{url}  
\usepackage{graphicx} 
\usepackage{natbib}  
\usepackage{caption} 
\DeclareCaptionStyle{ruled}{labelfont=normalfont,labelsep=colon,strut=off} 
\usepackage{algorithm}
\usepackage{algorithmic}

\usepackage{newfloat}
\usepackage{listings}
\floatstyle{ruled}
\newfloat{listing}{tb}{lst}{}
\floatname{listing}{Listing}

\usepackage{amsmath}
\usepackage{amssymb}
\usepackage{array}
\newcolumntype{P}[1]{>{\centering\arraybackslash}p{#1}}
\title{Discovering Spatial Relationships by Transformers for \\Domain Generalization}
\author{
    Cuicui Kang and 
    Karthik Nandakumar
}
\affiliations{
    cuicui.kang@mbzuai.ac.ae, karthik.nandakumar@mbzuai.ac.ae

}

\usepackage{bibentry}

\begin{document}

\maketitle
\begin{abstract}

Due to the rapid increase in the diversity of image data, the problem of domain generalization has received increased attention recently. While domain generalization is a challenging problem, it has been greatly advanced by deep architectures based on convolution neural nets (CNN). However, though CNNs have a strong ability for discriminative feature representation, they are not capable of modeling the relationships among different spatial locations of an image due to the lack of such operations in learnable filter based convolutions. On the other hand, the structural configurations of different local parts is very unique to a certain kind of object. These unique inner structures are very useful and important for characterizing an object as a whole, which are thought robust against domain shift. Considering this, this work tries to find the global feature structures by learning spatial relations of local parts, which tend to be consistent across domains and hence, improve the domain generalization ability. Specifically, upon a CNN backbone which is capable of both discriminant image feature representation, we explore the capability of Transformers to discover the feature structures between local CNN features. Accordingly, a hybrid architecture is developed, which is able to encode both discriminative local features and their global relationships to improve domain generalization. Evaluation on three well-known benchmarks demonstrates the benefits of modeling relationships between the features of an image using the proposed method and achieves state-of-the-art domain generalization performance. More specifically, the proposed algorithm outperforms the state-of-the-art by $2.2\%$ and $3.4\%$ on PACS and Office-Home databases, respectively.
\end{abstract}

\section{Introduction}
Many machine learning algorithms fail to work or experience a significant drop in their performance when they are applied on a dataset that is not encountered during training. Solving this domain shift problem is of critical importance in many applications. Domain Adaption (DA) \cite{Torralba-cvpy11-vlcs,Zhang-13icml,Zhao-nips18} and Domain Generalization (DG) \cite{Blanchard-nips11,zhao-nips20erm} are two approaches that have been proposed to address the problem of domain shift. In DA, samples of the target domain are provided without class information and the trained model is fine tuned based on these new samples to adapt the distribution. In contrast, DG aims to make the trained model work on unseen domain without providing any information about the target domain. Therefore, DG is considered to be a more challenging and generic problem and is the focus of this work.


\begin{figure}[t]
\begin{center}
   \includegraphics[width=0.95\linewidth]{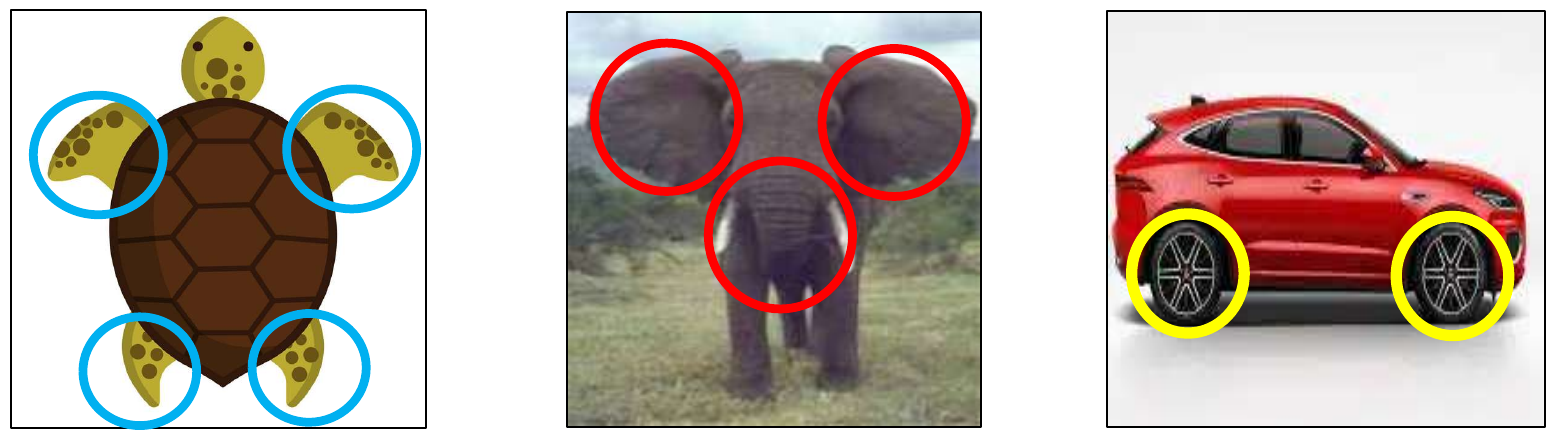}
\end{center}
\caption{The unique relationships between object parts for different objects.}
\label{fig:parts}
\end{figure}
During the past years, DG has become a well known problem and gained many attentions from both industry and academic. Many researches in various research fields have been sprung out to solve it from different views, leading great improvement on this topic.
Most of the existing methods are based on deep convolutional neural network (CNN). However, the CNN filters typically work on the local blocks of an image and ignore the relationships between the blocks. While, the parts connections and structure relations in a feature map is very unique to an object. For example, there are two similar wheels in a car on one side of the picture, as shown in Figure \ref{fig:parts}. These unique feature structures are very useful and important for characterizing an object as a whole. In order to capture long range dependencies for images, \cite{wangx-cvpr18nolocal} present non-local operations in CNN. Furthermore, \cite{wang-nips19globe} introduced patch-wise adversarial regularization to force the net focusing on the global concept of an object in DG problem.
With the global structures are relatively robust to domain shifts under consideration, this paper tries to learn the non local based global features with spatial relationships of local parts to improve the domain generalization ability.

\begin{figure}[t]
\begin{center}
   \includegraphics[width=0.99\linewidth]{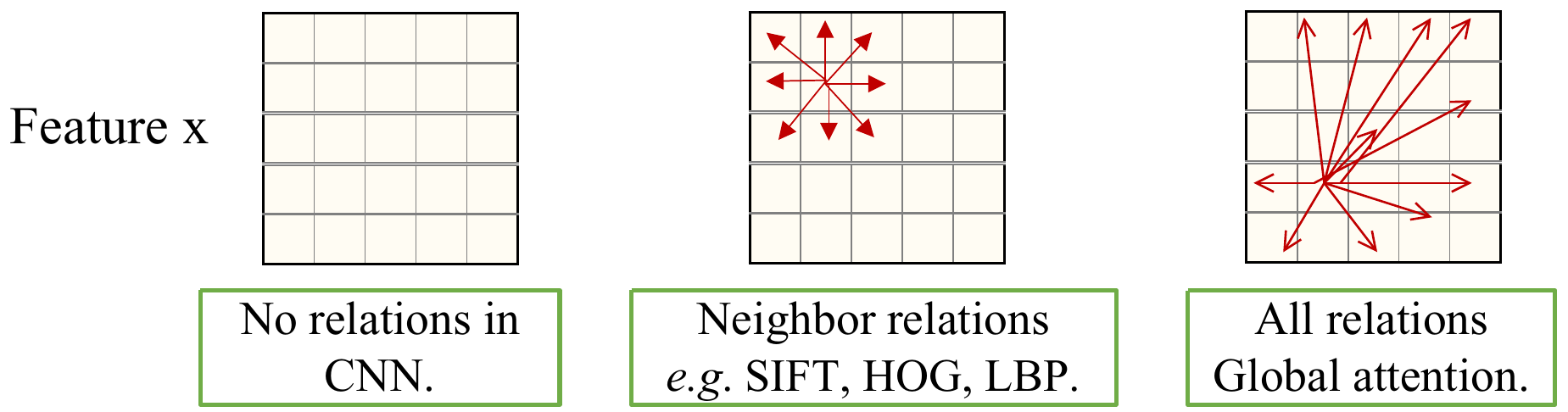}
\end{center}
   \caption{Illustration of relationship features in different feature extraction methods.}
\label{fig:short}
\end{figure}

Target at learning robust features that discover feature structures and encode parts relationships, we leverage the transformers framework for the spatial relationships modeling \cite{Vaswani-NIPS17,devlinl-2019-bert}, since CNN is not able to capture parts relationships in images. Suppose $\mathbf{x}$ is the feature, general convolution or pooling layer just uses $\mathbf{w}^T*\mathbf{x}$ to calculate the weighted sum, where $\mathbf{w}$ is the weight. But there is no inner comparisons in $\mathbf{x}$. The illustration of the differences between CNN, artificial features and self attention layer in Transformers can be found in Figure \ref{fig:short}. As shown in the figure, compared to CNN that doesn't utilize the spatial relationships, and some gradient based artificial features such as LBP and SIFT which have neighborhoods relations calculation, the Transformer computes global relationships no matter how far the features are. Actually, Transformer learns a matrix that encodes all the relations between the features.


%

Compared to existing DG algorithms, the primary novelty of this work is the global and generalized feature learning across domains with parts connections exploring. While Transformers achieved many successes in natural language processing, they are not so effective in extracting discriminative features for images compared to CNNs, except when an enormous amount of data is available for training \cite{alexey-2021vit}. And the pure Transformers often consume extensive computing resources. Therefore, we proposed a hybrid deep neural network architecture, which combines the benefits of both CNNs and Transformers to extract robust features with spatial connections that generalize well across domains. Specifically, we use a classic CNN based network as the backbone in the proposed architecture to ensure efficiency. Then, the convolutional features learned by the backbone are tokenized and forwarded to the Transformers for feature structure learning and spatial relationships discovering. Final classification is done with the output of Transformer features. The evaluation is done on three public and widely used data sets, and the comparisons to state-of-the-art algorithms show the proposed algorithm is robust and effective, which achieves the best performances, especially on the PACS and Office-Home databases where it outperforms the second best algorithms with clear improvements.

\section{Related Work}
\subsection{Existing Domain Generalization Methods}
The broad objective of DG is to utilize samples from multiple domains and learn a robust feature representation, which can generalize well to the unseen domain. In the recent past, many algorithms have been proposed for DG, which can be categorized to several categories based on their underlying motivations. One intuitive motivation is to learn the domain invariant feature representations or generic features across the domains. For instance, \cite{muandet-icml13} proposed a kernel-based optimization algorithm that learns an invariant transformation by minimizing the differences in the marginal distributions across the source domains. \cite{Ghifary-iccv2015dmtae} attempted to learn the domain-invariant features by the multi-domain reconstruction auto-encoders. \cite{Motiia-ICCV17} applied maximum mean discrepancy (MMD) to learn a latent space where the distance between images from different domains, but belonging to same category are minimized. \cite{Li-cvpr18mmdaae} also used MMD constraints with autoencoder to learn a domain-agnostic representation via adversarial training.  Recently, to ensure the conditional invariance of learned features, \cite{zhao-nips20erm} proposed an entropy-regularization approach that directly learns features that are invariant across domains. Besides, \cite{Matsuura-aaai2020} proposed to generate the pseudo domain labels by low layer features in a network, and then trained the domain-invariant feature extractor via adversarial learning.

Another class of DG algorithms, which also works at the feature level, employs model based approaches combining domain-agnostic and domain-specific parameters, and uses only the domain-agnostic features at the inference stage. This strategy was used in the shallow method by \cite{Khosla-eccv12} in the context of multi task learning. Then, \cite{lida-iccv2017deeper} developed a low-rank parameterized deep model for end-to-end domain generalization learning. \cite{Antonio-PR2018DSAM} introduced a deep architecture based on domain-specific aggregation modules, where generic perceptual information from multiple source domains are utilized. Recently, based on the instance normalization proposed by \cite{pan-eccv18IBN} and classic batch normalization, \cite{seo-eccv2020-dson} proposed to use multiple normalizations for each specific domain and then incorporate the optimized normalizations.

The third category of techniques employ various data augmentation strategies to improve the domain generalization ability. \cite{Shankar-iclr18} proposed a gradient-based domain perturbation strategy to perturb the input data with Bayesian Net. Using an adversarial strategy, \cite{volpi-nips18} augmented the data by synthesizing "hard" data in the training. \cite{Carlucci-CVPR19-jigen} proposed a self-supervised method with Jigsaw classifier and Jigsaw puzzle samples to capture more informative features for object classification, resulting in the well known JiGen algorithm \cite{Carlucci-CVPR19-jigen}. In order to deal with the unseen categories in unseen domain, \cite{Mancini-eccv20cumix} proposed to generate the images and features of unseen domain and unseen categories by mixing up the multiple source domains and categories used in the training.

Inspired by the Learning to Learn and Meta Learning paradigms, recent works have turned to optimization strategies for solving DG. For example, \cite{Li2018MLDG} proposed MLDG that trains on the split meta-train and meta-test sets on the source domains, which can be seen as the simulation of train/test domain shift in practice. \cite{Liya_2018_ECCV} also used meta learning for domain generalization, which simulated domain shift from the source domains during the training process. In \cite{Li-ICCV19Epilcr}, the authors designed an episodes training procedure for DG that exposed layers to neighbours that are untrained for the current domain. With episodic training procedure, \cite{Yogesh-nips2018metareg} used a regularization function in the classification layer to gain a general representation across domains. Finally, \cite{dou-nips2019masf} proposed a model-agnostic episodic learning procedure to regularize the semantic structure in the feature space.

This paper has a similar motivation to the work \cite{wang-nips19globe}, which argued that global structure plays a critical role in determining the class label and are more robust and general. They introduced patch-wise adversarial regularization to penalize the predictive power of local representations in the earlier layers of a neural network and employed the reverse gradient technique to extract global features instead of local features. In contrast, our approach directly attempts to model the spatial relationships between the local features using a transformer network and combine the predictive power of the local features and the robustness of the global features into a single representation.

\subsection{Transformers Related Works}

Transformer is an attention-based model that was firstly proposed by \cite{Vaswani-NIPS17}. It is a model architecture designed to learn global dependencies between input and output. The success of attention-based models in natural language processing inspired many researchers to exploit attention for solving problems in computer vision. For example, to deal with image classification, \cite{Woo-eccv18-cbam} developed the convolutional block attention module, \cite{wangx-cvpr18nolocal} presented the non-local blocks, \cite{Bello2019AttentionAC} suggested to concatenate the convolution feature and attention feature, and \cite{srinivas2021bottleneck} proposed Bottleneck Transformer with MHSA. Recently, \cite{alexey-2021vit} found that Transformer can be applied to image classification with raw image patches as input. The so-called vision transformer (ViT) was proposed and achieved performance levels comparable to state-of-the-art CNNs with enough training samples (e.g., JFT-300M with 300 millions images). However, training a ViT involves extensive computing resources and it cannot generalize well without sufficient training data. Therefore, \cite{touvron-2020deit} proposed a convolution-free transformer called DeiT that was trained on ImageNet only. This was achieved through token-based knowledge distillation using a CNN as a teacher. For object detection, \cite{Carion-eccv20} proposed DETR by using bipartite matching loss and transformers with parallel decoding. In contrast to these existing methods, the proposed approach leverages the Transformer encoder to learn the global feature structures that encode relationships between local parts for better DG ability.

\section{Proposed Method}

\begin{figure}[t]
\begin{center}
   \includegraphics[width=1\linewidth]{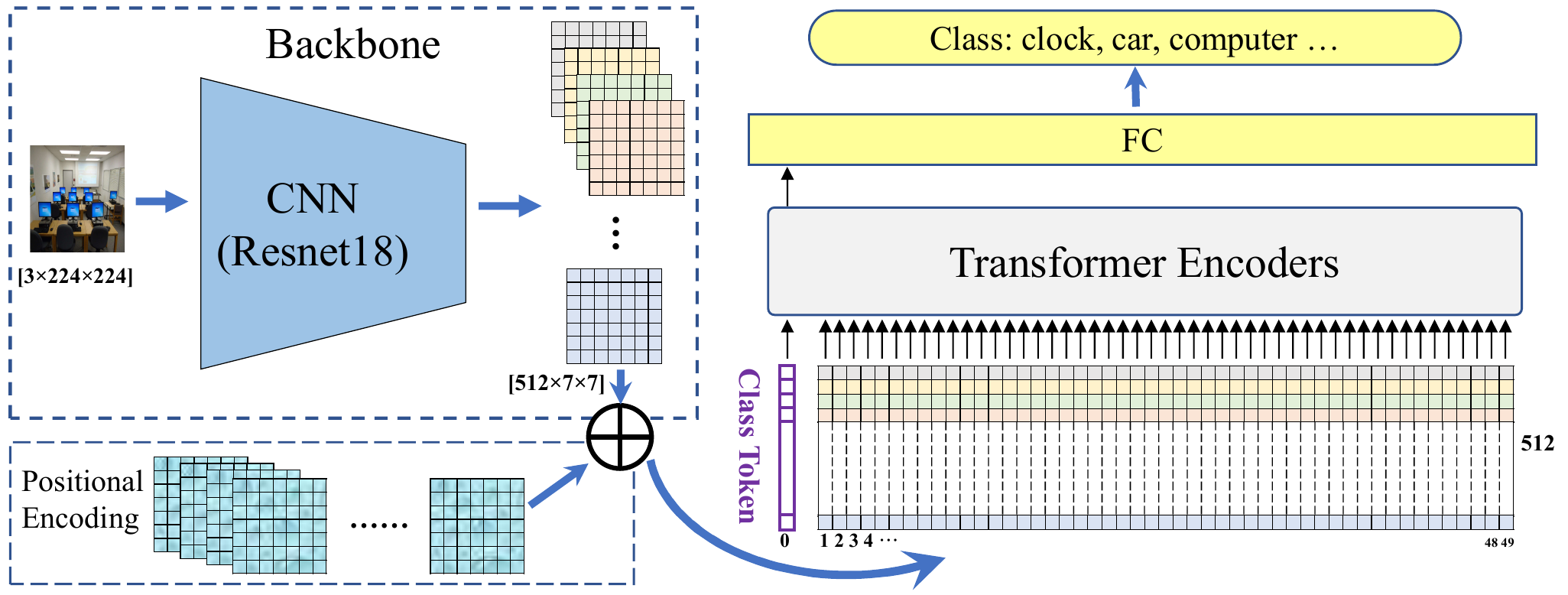}
\end{center}
   \caption{The architecture of the proposed ConvTran.}
\label{fig:arc}
\end{figure}

We propose a hybrid deep neural network architecture denoted as ConvTran to extract robust image representation that encapsulates both local discriminative features and spatial relationships between local parts. This is achieved by employing a regular CNN (e.g. ResNet) as the backbone and a Transformer encoder stack for spatial relationships modeling. The overall architecture of the proposed method is shown in Figure \ref{fig:arc}. The proposed approach is able to encode global structure and parts relationships for better domain generalization ability, and meanwhile the hybrid architecture avoids the need to train pure  transformers based on numerous images, thereby minimizing time, computational resources and data requirements.


\subsection{Transformer Formulation}

Before presenting the details of the proposed hybrid method, a brief introduction of the Transformer architecture is required. A classical Transformer \cite{Vaswani-NIPS17,devlinl-2019-bert} has two parts: the Encoder network and the Decoder network. In the proposed architecture, only the Transformer encoder stack is utilized to learn the spatial relationships. The Transformer encoder component consists of a stack of $L$ layers and each layer contains two main blocks, namely Multi-Head Attention (MHA) and Multi Layer Perceptron (MLP). In each block, Layer Normalization is used at the beginning to ensure and accelerate the convergence of the optimizer, as well as residual connections that are included at the end of each block.

The MHA is the core part of Transformers. It uses $h$ different linear projections to project the queries, keys, and values. The attention results are concatenated and once again projected to obtain the final representation. The basic attention model that is used in Transformers is the scaled dot-product attention, which is formulated as
\begin{equation}
\text{Attention}(Q, K, V) =\text{softmax} (\frac{QK^{T}}{\sqrt{d_k}})V,
\end{equation}
\noindent where $Q\in\mathbb{R}^{T\times d_k}$, $K\in\mathbb{R}^{M\times d_k}$ and $V\in\mathbb{R}^{M\times d_v}$ are the Query matrix, Key matrix and Value matrix, respectively. $d_k$ is the feature dimension of the key and query, and $d_v$ is the feature dimension of $V$. $T$ and $M$ are the sequence lengths of the query and key/value.

Based on the scale dot-product attention, Transformer combines it with multi-head attention, which is defined as:
\begin{equation}
\text{Multi-Head}(Q, K, V) =\text{Concat}(H_1 , H_2 , ... , H_h )W^O, \nonumber
\end{equation}
\begin{equation}
\text{where~}H_i =\text{Attention}(QW_i^Q, KW_i^K, VW_i^V),
\end{equation}
where $W_i^Q\in\mathbb{R}^{d\times d_k}$, $W_i^K\in\mathbb{R}^{d\times d_k}$ and $W_i^V\in\mathbb{R}^{d\times d_v}$ are learned projection matrices for head $H_i$. $W_O\in\mathbb{R}^{hd_v\times d}$ is also a parameter matrix that is multiplied to the concatenation of $h$ heads attention results to get the final output. Note that, Transformer encoder uses the multi-heads self attention model (MHSA), which means that $Q=K=V$.

At the end of MHA block, residual addition is performed before the features are passed on to MLP module. The MLP is a position-wise fully connected feed-forward network that contains two linear layers with a RELU non-linearity in between.

Let $\mathbf{x}_{\ell}$ denote the input to the $\ell^{th}$ transformer encoder layer. The forwarding functions of the $\ell^{th}$ layer can be summarized as:
%
\begin{align}\label{eq:trans}
& \mathbf{x}'_\ell = MHA(LN(\mathbf{x}_{\ell} )) +\mathbf{x}_{\ell}, &\ell\in[0, 1, ... , L-1], \\
& \mathbf{x}_{\ell+1} = MLP(LN(\mathbf{x}'_{\ell} ))+\mathbf{x}'_{\ell}, &\ell\in[0, 1, ... , L-1],
\end{align}
\noindent where $LN$ denotes the Layer Normalization function.

\subsection{The Proposed Hybrid Architecture}
In the proposed approach, a convolution neural network is used as the backbone, and the extracted convolutional features serve as the input of the Transformer encoders as shown in Figure \ref{fig:arc}. Since the traditional Transformers receive sequence of feature embeddings as input, we reshape the extracted features by CNN network which are original usually stacked in block, to matrix, and view them as a sequence of tokens to learn the parts spatial relationships. In other words, the convolutional features extracted by CNN filters are treated as words constituting the sentence (whole image). Suppose that $\mathbf{x}_{c}\in\mathbb{R}^{P\times P\times C_f}$ is the corresponding convolutional features with $C_f$ channels. The features are then reshaped to $\mathbf{x}_{f}\in\mathbb{R}^{K\times C_f}$ before they are fed to Transformer encoders, where $K=P\times P$. This can be considered as a sequence of $K$ words, where each word is defined by a $C_f$-dimensional descriptor.
%


Transformer includes positional encodings to retain the knowledge about the order of inputs. In our implementation, learnable 1-D position embeddings are used. In practice, we found that the prior positional encodings with sine and cosine functions can also work as well as the learnable positional encodings. Moreover, the class token is also utilized in a similar way to BERT \cite{devlinl-2019-bert}, so as to aggregate sequence representations for classification tasks. Thus, the input to the first transformer layer is:
%
\begin{equation}
\mathbf{x}_0 = [\mathbf{x}_{token};\mathbf{x}_{f}] + E_{pos},
\end{equation}
\noindent where $\mathbf{x}_0, E_{pos}\in \mathbb{R}^{D \times C}$, $D = (K+1)$ and $C=C_f$. Here, $\mathbf{x}_{class} \in \mathbb{R}^{1 \times C}$ is the class token, $\mathbf{x}_{f}$ is the output of the CNN, and $E_{pos}$ is the 1-D standard learnable positional encodings. After learning the transformer encoder layers, only the learned classification token is sent to the final fully connected layer for further classification.

\textbf{ResNet based ConvTran}: Considering the success of residual connection networks, ResNet \cite{he-resnet} is a good choice for the backbone CNN in the proposed architecture. We denote this specific implementation as our primary model. In ConvTran, the outputs of the last convolution layer of ResNet without pooling process are forwarded to the Transformer encoders. We also incorporate Instance Normalization (IBN-B) into the ResNet framework due to its improvement to the generalization ability according to researches \cite{pan-eccv18IBN,seo-eccv2020-dson}.

\section{Experiments}



\subsection{Implementation Details}

For the experiments, the proposed architecture is first pre-trained in an end-to-end fashion on the ImageNet dataset by using a SGD optimizer. The pre-trained model is then fine-tuned on the training data of the experimental dataset and the resulting model is applied to the test set. The output features of Layer4 in a ResNet are fed to the Transformer encoders, which consist of $L=2$ layers. For example, the Layer4 output features in ResNet18 have 512 channels, each representing of $7\times 7$ filter results. Thus, the input features to Transformer encoders have a dimension of $50\times 512$ after inserting with the class token in the first row. The number of heads in the MHSA module is two and the dimension of feed forward layer in the MLP module is set to 1024. Further discussions about transformer parameters can be found in the ablation study afterwards. For fine-tuning on the experimental datasets, the SGD algorithm with learning rate of 0.001 is used. The maximum of epochs is set to 100. All reported accuracy values are based on an average of ten runs for every test scenario.


\begin{table*}
\begin{center}
\begin{tabular}{|l@{~}|P{0.33in}|P{0.35in}|P{0.33in}|P{0.35in}|P{0.37in}|P{0.35in}|P{0.35in}|P{0.35in}|P{0.35in}|P{0.32in}|P{0.35in}|P{0.44in}|}
\hline
\small{Methods}        & AGG  & DSAM   & JiGen   & MLDG   & Metareg  & MASF   & Epifcr  & Cumix & MMLD &ER
&DADG & \small{ConvTran} \\
\hline\hline
Photo           & 94.40 & 95.30 & 96.03  & 94.00  & 95.50   & 94.99 & 93.90 & 95.10  & 96.09 & \textbf{96.65}& 94.86  &96.22   \\
\hline
Art             & 77.60 & 77.33 & 79.42  & 78.70  & 83.70   & 80.29 & 82.10 & 82.30  & 81.28 & 80.70  & 79.89  & \textbf{84.04}  \\
\hline
\small{Cartoon} & 73.90 & 75.89 & 75.25  & 73.30  & 77.20   & 71.17 & 77.00 & 76.50  & 77.16 & 76.40  & 76.25  & \textbf{78.98}  \\
\hline
Sketch          & 70.30 & 69.27 & 71.35  & 65.10  & 70.30   & 71.69 & 73.00 & 72.60  & 72.29 & 71.77  & 70.51  &  \textbf{76.99}  \\
\hline\hline
Ave.            & 79.10 & 80.72 & 80.51  & 80.70  & 81.70   & 81.03 & 81.50 & 81.60  & 81.83 & 81.38  & 80.38  &  \textbf{84.05}  \\
\hline
\end{tabular}
\end{center}
\caption{The results on PACS dataset for the proposed ConvTran (Resnet18 as backbone) and the compared algorithms.}
\label{tlb:pacs-res18}
\end{table*}

\subsection{Datasets}
\textbf{PACS} dataset, as the most famous and popularly experimented dataset~\cite{lida-iccv2017deeper} in DG field, contains 9991 images coming from seven categories and four domains, which are Photo (P), Art Painting (A), Cartoon (C) and Sketches (S). It can be downloaded free for research purpose\footnote{\url{https://domaingeneralization.github.io/#data}}, and all the pictures are in size of 227$\times$227 for convenience. For fair comparison, we follow the experimental protocol in \cite{lida-iccv2017deeper} to train on the training sets of three domains and test on the remaining domain. \textbf{VLCS} is another popular domain generalization evaluation dataset \cite{Torralba-cvpy11-vlcs}. It consists of 5 categories shared by PASCAL VOC 2007, Labelme, CALTECH and SUN datasets, which in turn act as the four domains. There are 10729 images in the resolution of 227$\times$227. The experimental protocol from \cite{Ghifary-iccv2015dmtae} is used that each domain in the VLCS dataset was divided into a training set (70\%) and a test set (30\%) by random selection from the overall data set, and leave-one-domain-out for test protocol is also followed in the experiment. \textbf{Office-Home} dataset contains 65 categories of daily used objects from four domains, which are Art, Clipart, Product and Real World. It was originally proposed in the Domain Adaption field for object recognition~\cite{venka-cvpr17-office}. Recently, it was applied to support the DG problem and has got widely evaluated in short time due to its variety in categories \cite{Antonio-PR2018DSAM, Carlucci-CVPR19-jigen, huang-eccv20rsc,seo-eccv2020-dson}. We also followed the same experimental protocol as on PACS dataset as introduced in \cite{lida-iccv2017deeper}. 

\begin{table}[t]
\begin{center}
\begin{tabular}{|l|P{0.33in}|P{0.3in}|P{0.33in}|P{0.33in}|P{0.44in}|}
\hline
\small{Methods}        & DSAM   &   JiGen  &   MLDG   & DADG   & \small{ConvTran}  \\
\hline\hline
Art           & 58.03 & 53.04  & 52.88   & 55.57  & \textbf{60.35}    \\ \hline
Clipart       & 44.37 & 47.51  & 45.72   & 48.71  & \textbf{52.84}    \\ \hline
Product       & 69.22 & 71.47  & 69.90   & 70.90  & \textbf{73.67}    \\ \hline
RWorld        & 71.45 & 72.79  & 72.68   & 73.70  & \textbf{75.87}    \\ \hline\hline
Ave.          & 60.77 & 61.20  & 60.30   & 62.22  & \textbf{65.68}    \\
\hline
\end{tabular}
\end{center}
\caption{The results on Office-Home for the proposed ConvTran (Resnet18) and the compared algorithms.}
\label{tlb:office-res18}
\end{table}

\begin{table*}[t]
\begin{center}
\begin{tabular}{|m{0.46in}@{~}|P{0.50in}|P{0.35in}|P{0.39in}|P{0.39in}|P{0.39in}|P{0.39in}|P{0.39in}|P{0.39in}|P{0.32in}|P{0.39in}|P{0.45in}|}
\hline
\small{Methods}        & \small{MMDAAE}  & AGG   & DSAM  & JiGen    & MLDG     & Epifcr   & MASF      & MMLD &ER & DADG & \small{ConvTran}    \\
\hline\hline
Caltech         & 94.40   & 93.10  & 91.75  & 96.93  & 94.40 & 94.10 & 94.78  & 96.66 & \textbf{96.92} & 96.80  & 96.58 \\ \hline
\small{Labelme} & 62.60   & 60.60  & 56.95  & 60.90  & 61.30 & 64.30 & 64.90  & 58.77 & 58.26  & \textbf{66.81} & 66.57  \\ \hline
Pascal          & 67.70   & 65.40  & 58.59  & 70.62  & 67.70 & 67.10 & 69.14  & 71.96 & \textbf{73.24} & 70.77  & 70.05  \\ \hline
Sun             & 64.40   & 65.80  & 60.84  & 64.30  & 65.90 & 65.90 & 67.64  & 68.13 & \textbf{69.10} & 63.64  & 65.62  \\ \hline \hline
Ave.            & 72.30   & 71.20  & 67.03  & 73.19  & 72.30 & 72.90 & 74.11  & 73.88 & 74.38  & 74.46          & \textbf{74.71}\\
\hline
\end{tabular}
\end{center}
\caption{The experimental results on VLCS for the proposed ConvTran with Alexnet as backbone.  }
\label{tlb:vlcs-alex}
\end{table*}

\subsection{Comparison Results to State-of-the-art}
To evaluate the proposed method, we compare it to many state-of-the-art algorithms such as MMD-AAE \cite{Li-cvpr18mmdaae}, MLDG \cite{Li2018MLDG}, MetaReg \cite{Yogesh-nips2018metareg}, MASF \cite{dou-nips2019masf}, JiGen \cite{Carlucci-CVPR19-jigen}, Epi-fcr, \cite{Li-ICCV19Epilcr}, DADG \cite{Chen-2020dadg}, MMLD \cite{Matsuura-aaai2020}, D-SAM \cite{Antonio-PR2018DSAM}, Cumix \cite{Mancini-eccv20cumix} and ER \cite{zhao-nips20erm}.

\textbf{PACS dataset}. The experimental results on the dataset are shown in Table \ref{tlb:pacs-res18}, where the best performances are in bold for each domain task. Note that, the results of D-SAM, JiGen, Cumix, MASF, Metareg, AGG, MMLD, ER and Epi-fcr algorithms are taken from the referenced paper of the corresponding author, except for the MLDG method, which is from the first author's another paper \cite{Li-ICCV19Epilcr}. From the comparison, we can see that the proposed algorithm outperforms the second best algorithm for more than 2.2\% on average, which achieves 84.05\% compared to the second one with 81.83\% by MMLD. The following performances are 81.70\% by MetaReg, 81.60\% by Cumix, and 81.50\% by Epi-fcr, that are very close to each other. On the separate tasks with different target domains, the proposed algorithm gets the best performances on Art-painting, Cartoon and Sketch domains, and very close performance to the best on the original Photo domain (96.22\% by us vs. 96.65\% by ER). Especially on Cartoon and Sketch domains, the proposed ConvTran achieves 78.98\% and 76.99\%. Compared to the second best performance that are 77.20\% by MetaReg on cartoon domain and 73.0\% by Epi-fcr on sketch domain, it improves 1.78\% and 3.99\% respectively. With the clear gap we can see that utilizing global feature structure indeed benefits the domain generalization ability and reduces the domain shift effects.

\textbf{Office-Home} is a challenging dataset in DG field, especially for the $Clipart$ domain. The experimental results are displayed in Table \ref{tlb:office-res18}, where the performances of D-SAM, JiGen and DADG are taken from the referenced paper, and the MLDG is taken from \cite{Chen-2020dadg}. From the table, it can be clearly seen that the proposed algorithm gets 65.68\% for average accuracy of four domains, which achieves the best result and outperforms the second best algorithm for more than 3\% (compared to 62.22\% by DADG). On each sub-task, the ConvTran outperforms the second best method for more than 2\% (60.35\% vs. 58.03\% by D-SAM) on Art-Painting, for more than 4\% to the second DADG on Clipart, more than 2\% on Product domain (compared to 71.47\% by JiGen), and 2\% on Real World domain (compared to 73.70\% by DADG). Similar to the PACS data, the proposed algorithm gets the SOTA performance on each domain test, and shows its great domain generalization ability with the discovered global structure.

The results on \textbf{VLCS} dataset are shown in Table \ref{tlb:vlcs-alex}. Due to some reasons, most of previous works are tested with Caffe pretrained Alex net only on the VLCS dataset. Considering this, we also evaluate the proposed ConvTran with Alexnet as backbone on it for fair comparison. Note that the performance of D-SAM is taken from \cite{Carlucci-CVPR19-jigen} where the owner of D-SAM is a co-author. From the Table, we can see that the proposed ConvTran algorithm is comparable to the state-of-the-art algorithms. It reaches 74.71\% accuracy on average, which is comparable to the second best algorithm DADG with 74.46\% and third algorithm ER with 74.38\%. It also outperforms the famous algorithms JiGen and MLDG with improvements of 1.5\% and 2.4\%, respectively. Compared to ER, the ConvTran is comparable on the domains of Caltech, Pascal Voc and Sun. However, the ER fails to generalize well on Labelme domain, only gets 58.26\%. In contrast, the ConvTran reaches 66.57\%, which is very close to the best result 66.81\% by DADG. Without bias on the various domains, the proposed architecture achieves the best overall algorithm on VLCS dataset, which is comparable to the SOTA and indicates its robust domain generalization ability.

\begin{table}[t]
\begin{center}
\begin{tabular}{|l|c|c|c|c|c|}
\hline
\multicolumn{6}{|c|}{\textit{PACS data set}}     \\ \hline
& P  & A     & C  & S & Avg. \\\hline
\small{Baseline-18}    &     94.47 &     78.02 &     75.92  &     75.69 &     81.02   \\ \hline
\small{IBN-b-18}       &     94.19 &     80.25 &     77.64  &     76.33 &     82.10   \\ \hline
\small{\bf ConvTran-18} & \bf 96.22 & \bf 84.04 & \bf 78.98  & \bf 76.99 & \bf 84.05  \\ \hline
\small{Baseline-50}    &     96.84 &     85.50 &     78.88  &     73.04 &     83.57 \\ \hline
\small{IBN-b-50}       &     97.44 &     85.17 &     82.39  &     77.04 &     85.26  \\ \hline
\small{\bf ConvTran-50} & \bf 97.92 & \bf 87.12 & \bf 83.78  & \bf 78.00 & \bf 86.71   \\ \hline
\multicolumn{6}{|c|}{\textit{Office-Home data set}}     \\ \hline
            & A   &  C  & P & R &  Avg. \\ \hline
\small{Baseline-18}    &     53.02  &     50.02  &     71.25  &     72.92  &     61.80   \\ \hline
\small{IBN-b-18}       &     54.49  &     51.44  &     71.18  &     73.43  &     62.64   \\ \hline
\small{\bf ConvTran-18} & \bf 60.35  & \bf 52.84  & \bf 73.67  & \bf 75.87  & \bf 65.68   \\  \hline
\small{Baseline-50}    &     63.92  &     54.11  &     77.97  &     79.76  &     68.94 \\ \hline
\small{IBN-b-50}       &     66.07  &     55.94  &     77.19  &     79.86  &     69.76   \\ \hline
\small{\bf ConvTran-50} & \bf 69.95  & \bf 57.28  & \bf 80.18  & \bf 82.50  & \bf 72.48   \\  \hline
\end{tabular}
\end{center}
\caption{The ablation study of the proposed ConvTran with ResNet18 and ResNet50 as backbone on PACS and Office-Home datasets.}
\label{tlb:abla-base18}
\end{table}

\subsection{Ablation Study}
\textbf{Comparison to Different Networks}

The proposed ConvTran is compared to the ResNet architecture for ablation study, where both ResNet18 and ResNet50 are taken as the baselines. We also include the Resnet-ibnb network in this comparison, abbreviated as IBN-b. The comparisons are conducted on the PACS and Office-Home datasets, and the results are shown in Table \ref{tlb:abla-base18}.

For PACS dataset, we can see that the overall average accuracy of the baseline algorithm is 81.02\%. With adding of the instance normalization, the performance reaches to 81.20\% by Resnet18-ibnb. As for the proposed ConvTran, it achieved 84.05\%, with almost 2\% improvement to IBN-b. It obvious that the proposed ConvTran improves more performance to IBN-b than improvement for IBN-B to baseline. With Resnet50 as backbone, the proposed algorithm improves almost 1.5\% to IBN-b model, which is comparable to the improvement about 1.7\% from IBN-b to the baseline algorithm.
As for Office-Home dataset with Resnet18 architecture, the overall average performance of the proposed ConvTran outperforms IBN-b for 3.04\%. However, the IBN-b only gets 0.8\% improvement compared to the baseline algorithm. From the comparison, we can see the proposed ConvTran benefits more domain generalization than IBN-b method. As for Resnet50 for backbone, the proposed algorithm outperforms the IBN-b for 2.7\% on average, while the IBN-b gets only 0.8\% compared to the baseline.
Through the comparison we can see that, the proposed algorithm achieves more improvements by utilizing the global structures than Instance Normalization to the generalization ability across domains. The ablation study results with comparison to the baseline and IBN-b architecture indicate the proposed hybrid architecture engaged to learn the general and robust features across the multiple domains with Transformer importation. The performances on both PACS and Office-Home datasets prove that the global structure information captured by ConvTran with spatial relationships helps to improve the domain generalization ability.


%

\begin{table}[t]
\begin{center}
\begin{tabular}{|l|c|c|c|c|}
\hline
LayerNum    & 2 layers  &  4 layers &   6 layers &  8 layers\\
\hline\hline
Photo      & 95.11	&95.11	&94.78	&94.77    \\ \hline
Art        & 81.56	&80.18	&80.61	&80.29    \\ \hline
Cartoon    & 78.18	&77.18	&76.90	&76.89    \\ \hline
Sketch     & 76.22	&75.53	&74.15	&74.35    \\ \hline\hline
Average    & 82.77	&82.00	&81.61	&81.57    \\
\hline
\end{tabular}
\end{center}
\caption{The ablation study of transformers' encoder layers in the proposed ConvTran (Resnet18) on PACS dataset.}
\label{tlb:abla-layer}
\end{table}

\textbf{Why small Transformer in the proposed method}

In this part, we study the hyper parameters' effect of Transformer encoders to the performance. For convenience, we used the IBN-b pretrained model in the proposed algorithm. Note that PACS dataset is used in this part and each test has been ran ten times to get the final performance on average. Firstly, the proposed algorithm with two, four, six and eight transformer layers are evaluated. The results can be found in Table \ref{tlb:abla-layer}. From the results, it can be seen there is gradually and slowly performance descending against the depth of the Transformers. Compared the model with two layers and eight layers, the shallow one is 1\% higher than the deep one. But the performances are very close between the neighbours. One of the reasons may be that the Transformers need sufficient training samples to show its power, but most of the well known DG datasets are small datasets. Larger models need more pre-training samples to support, and more computation resources and more times for training. We did not choose a large model since the benchmarks in DG are small. Besides, the proposed algorithm is a hybrid deep net with ResNet, which also supports us for utilizing a small transformer encoding net. However, according to the researches of \cite{alexey-2021vit,touvron-2020deit}, deep model can be used when there are more data supports, which will serve as follow-up research.

Then, the heads number in MHSA block is also analyzed, and the results are displayed in Table \ref{tlb:abla-head}. From the table we can see that the performances with heads of 2, 4, 8 and 16 are very close to each other. As well as the performances with different dimensions of the feed forwarding features in MLP block, which has been shown in Table \ref{tlb:abla-dim}. From the Table, it can be seen that there is no obvious performance difference between 512, 1024, 2048 and 4096 of the feature dimensions on average, which is similar to the parameter study on multiple heads. The reason may lays in that small benchmark datasets in DG field do not need large Transformer network for modeling. The large network of Transformers needs more data to ensure the performance. 

\begin{table}[t]
\begin{center}
\begin{tabular}{|l|c|c|c|c|}
\hline
HeadNum     & 2 heads  &  4 heads &   8 heads &  16 heads\\
\hline\hline
Photo       & 95.11	&94.93	&95.02	&94.90     \\ \hline
Art         & 81.56	&81.31	&81.44	&81.00     \\ \hline
Cartoon     & 78.18	&78.68	&78.34	&78.35     \\ \hline
Sketch      & 76.22	&75.92	&75.32	&76.20     \\ \hline\hline
Average     & 82.77	&82.71	&82.53	&82.61    \\
\hline
\end{tabular}
\end{center}
\caption{The ablation study of transformers' Multi Heads number in the proposed ConvTran (Resnet18) on PACS.}
\label{tlb:abla-head}
\end{table}

\begin{table}[t]
\begin{center}
\begin{tabular}{|l|c|c|c|c|}
\hline
Dimension     & 512-D &  1024-D &  2048-D & 4096-D\\
\hline\hline
Photo          & 94.87	& 95.11	& 95.23	& 95.05     \\ \hline
Art            & 81.78	& 81.56	& 80.97	& 81.44     \\ \hline
Cartoon        & 78.10	& 78.18	& 77.87	& 77.86     \\ \hline
Sketch         & 75.72	& 76.22	& 76.43	& 75.67     \\ \hline\hline
Average        & 82.62	& 82.77	& 82.62	& 82.50    \\
\hline
\end{tabular}
\end{center}
\caption{The ablation study of transformers feed forward dimensions in the proposed ConvTran (Resnet18) on PACS.}
\label{tlb:abla-dim}
\end{table}

To summarize, it can be seen that the proposed algorithm is not sensitive to the hyper parameters. Besides, the results with different depth of transformers indicate that deep transformers network needs more sufficient samples for training to show its power. However, small transformers in hybrid network is a better choice with efficient performances and less computation resources occupied, if dataset is not huge.



\textbf{Insights Into Learned Attention Maps}		

\begin{figure}[t]
\begin{center}
   \includegraphics[width=0.99\linewidth]{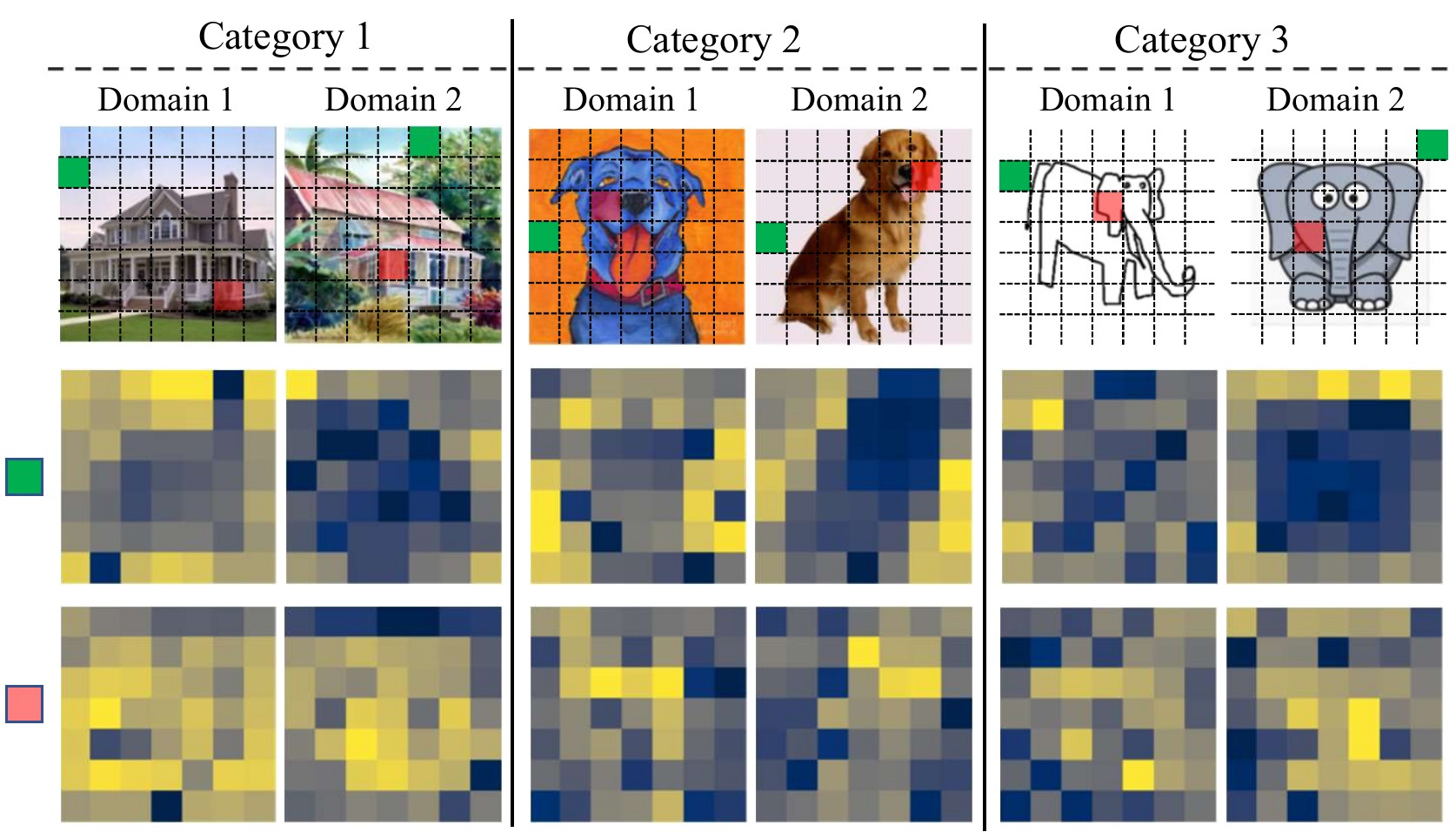}
\end{center}
   \caption{Examples of the learned attention matrix for the samples from different domains. The 1st row contains original images with randomly selected locations, 2nd row images are the attention maps to block (green) from the background and 3rd row images are the attention maps to the block (red) from the object.}
\label{fig:visual}
\end{figure}

With the questions how Transformers learn the parts relationships between different spatial locations in an image and do they really get the relationships, we take an inspection of the MHSA layers in Transformers. The learned attention maps in each transformer layer are matrices in dimension of $49\times49$ excluding the class token in the proposed ConvTran. Thus, we randomly choose two blocks in the image, which belong to the background and target object respectively. Then we take out the corresponding vector in the attention matrix and reshape it to $7\times 7$ for viewing. There are some learned attention/relation maps are shown in Figure \ref{fig:visual}, where the sample images located in the first row are randomly selected from different domains on PACS. In Figure \ref{fig:visual}, the second row shows the attention maps to the block in background, and the third row contains the attention maps to the block belonging to object. The darker blocks in the attention maps indicate less relevant, while the lighter blocks are more relevant.

From the figure, it can be seen that the proposed ConvTran can learn the relationships between the blocks that are even far away from each other. The block from background is more relevant to the local blocks containing background, and less relevant to the blocks from the object usually. The block from object often has stronger relevance to the local blocks also contain object parts, and sometimes has high relationship with block from the background. But the orientation of the attention maps is consistent with the input image on the overall. With different background locations from different domains, the ConvTran can also find the related background blocks. As well as the objects, even when the object parts' locations are different in images, the related object parts are still discovered despite of domain shift. For example, for category 2 in Figure \ref{fig:visual}, the red blocks from domain 1 and domain 2 images are both on the dogs' face. We can see that the proposed ConvTran successfully found the highly related object parts are mostly from the dog's face for both domains. It is obvious that the global structure is successfully discovered by ConvTran, which is robust to distortions of objects and less effected by domain shift. From the figure we can see that parts relationships can be learned even when the parts’ spatial locations are changed, and the global structure relationships which are robust to distorted images benefit the domain generalization ability finally.


\section{Conclusion}

To deal with domain generalization problem, we attempt to learn generalized features with spatial relationship that encode global feature structures. Therefore, we proposed to connect the attention model, namely Transformer, to a CNN based deep network, namely ResNet. We also demonstrated how the global parts relationships learned by Transformers benefited domain generalization. The experiments are conducted on three popular and widely used databases with comparison to the state-of-the-art algorithms, showing that the proposed algorithm achieves the best performance. Through the ablation study, it can be seen that the proposed algorithm is robust and does learn global structures in features that improve generalization ability across domains.


\appendix

\bibliography{egbib}

\end{document}